\documentclass[11pt,a4paper]{article}

\usepackage[utf8]{inputenc}
\usepackage[T1]{fontenc}
\usepackage{amsmath,amssymb,amsfonts}
\usepackage{graphicx}
\usepackage{booktabs}
\usepackage{hyperref}
\hypersetup{%
  pdftitle={Ontology-Amplified Distillation and Contextuality Auditing for Sovereign Enterprise Language Models},
  pdfauthor={Thanh Luong Tuan},
  pdfkeywords={ontology-amplified distillation, sovereign language models, enterprise AI, contextuality audit, direct preference optimization},
  colorlinks=true,
  linkcolor=black,
  citecolor=black,
  urlcolor=blue
}
\usepackage{cleveref}
\usepackage[margin=1in]{geometry}
\usepackage{natbib}
\usepackage{enumitem}
\usepackage{multirow}
\usepackage{amsthm}
\newtheorem{definition}{Definition}

\newcommand{\faos}{\textsc{FAOS}}
\newcommand{\ronto}{r_{\mathrm{onto}}}

\title{%
  Ontology-Amplified Distillation and Contextuality Auditing\\
  for Sovereign Enterprise Language Models:\\
  A Combined Proof-of-Mechanism and Negative-Results Method Study%
}

\author{%
  Thanh Luong Tuan%
  \thanks{Corresponding author. Email: tluongtuan@my.ggu.edu.
    ORCID: 0009-0000-1199-837X.} \\
  \textit{Foundation AgenticOS (\faos{})}%
}

\date{July 2026}

\begin{document}
\maketitle

\begin{abstract}
Regulated financial institutions operating under data-residency rules need
tenant-owned language models that can run inside the institution's perimeter.
This paper combines two related FAOS studies into one mechanism-and-control
article. First, it reports a reduced-power proof-of-mechanism study of
\emph{ontology-amplified distillation}: a Qwen3.6-27B student is adapted to the
Foundation AgenticOS ontology through supervised fine-tuning on frontier-teacher
trajectories and ontology-grounded direct preference optimization (DPO), trained
locally on a single Apple M5~Max from 47 synthetic, English-language,
cross-domain preference pairs. On 40 held-out Vietnamese financial-domain tasks,
the distilled student grounds 36 of 40 tasks (grounded rate 0.90; mean ontology
term-coverage $r_{\mathrm{onto}}=0.95$ on a metric floored at 0.50), equal to the
GPT-5 frontier baseline, which also grounds 36 of 40. The outcome is
underpowered to establish equivalence: the paired-difference 95\% confidence
interval spans $\pm4$ tasks, and the run does not test or show the
pre-registered amplification prediction that the student should exceed the
frontier. Second, the paper consolidates a contextuality-audit method for
enterprise-agent routing. In a separate negative-results pilot, the corrected
canonical Contextuality-by-Default degree is zero for all Phase~1.3 groups in
both the local-Qwen run and an explicitly labeled Gemma replication check; the
useful signal is direct influence and construct coupling, not surviving residual
contextuality. Together, the studies pair an ontology-grounded model-building
mechanism with a governance diagnostic for deciding when apparent disagreement
should trigger prompt standardization, multi-agent synthesis, or human review.
The evidence supports neither deployability, safety, superiority, statistical
equivalence, nor a contextuality-positive routing rule.
\end{abstract}

\section{Introduction}
\label{sec:intro}

Financial institutions in Vietnam operate under data-residency and
sector-specific rules that restrict where customer data may be processed
\citep{vnailaw2025, vnfintechsandbox2025}. For a regulated tenant, sending
account records or underwriting files to a foreign frontier-model API is not a
procurement decision but a compliance violation. A model that runs inside the
institution's perimeter therefore becomes a deployment requirement, and the
frontier model functions as a quality ceiling rather than a deployment
alternative. This framing reverses the usual cost--quality trade-off. The
question is not whether a smaller model is cheaper to run, but whether a locally
deployable model can reach the grounding quality that regulated work demands.

Earlier work established that supplying a domain ontology to a frontier model
improves its domain fidelity \citep{luong2026neurosymbolic, sharma2025ograg,
liu2025ontotune}. That work measured the effect on a frontier model. It left
open the question that matters for sovereign deployment: whether the same
ontology grounding can lift a smaller, locally deployable student to the
frontier's level on regulated-domain tasks. A distilled student carries thinner
parametric coverage of a specialized domain than a frontier model does, so
ontology grounding may matter more for the student---a larger relative gain from
a lower base---without necessarily closing the absolute gap on the hardest
tasks. The companion neurosymbolic study named this pattern the Inverse
Parametric Knowledge Effect (Inverse PKE): grounding value rises as the model's
parametric coverage of a domain falls \citep{luong2026neurosymbolic}.

This paper studies that question through \emph{ontology-amplified distillation},
a two-stage adaptation. A student model is first supervised-fine-tuned on
frontier-teacher trajectories that carry an injected ontology slice, then tuned
with ontology-grounded direct preference optimization
\citep{rafailov2023dpo, hinton2015distillation}. The preference signal pairs an
ontology-grounded frontier answer, marked preferred, against an ungrounded
base-model answer, marked dispreferred, so the student learns to reproduce
ontology-grounded behavior rather than to imitate the frontier's full parametric
knowledge. The student is a Qwen3.6-27B model \citep{qwen2025qwen25}; training
and evaluation run locally in Apple's MLX framework on a single M5~Max
\citep{applemlxm5}, and the trained student is fused to a four-bit checkpoint of
roughly 14~GB for deployment-consistent inference.

On a held-out set of 40 Vietnamese financial-domain tasks, the distilled student
grounds 36 of 40 tasks---a grounded rate of 0.90, or mean term-coverage
$r_{\mathrm{onto}} = 0.95$ on a metric floored at 0.50---equal to the GPT-5
frontier baseline, which also grounds 36 of 40. The $n=40$ binary outcome cannot
establish equivalence: the paired-difference 95\% confidence interval spans
$\pm 4$ tasks. Training-side measurements are consistent with the intended
preference shift: preference accuracy moves from 0 to 1.0 and the reward margin
from 0 to 0.307 over a single epoch, on a five-pair validation split. The result
is scoped as a reduced-power proof of mechanism, not a deployment claim, and not
the pre-registered amplification---which predicted the student \emph{exceeding},
not equalling, the frontier. One binding gate was run---ontology compliance---on a
metric that records whether the answer surfaces at least one of the task's
ontology terms, a degenerate proxy for the registered four-component composite,
not whether the answer is complete or correct. The preference data are synthetic,
English-only, and drawn from non-target domains; the frontier ran at minimal
reasoning effort; the student matches rather than exceeds the frontier; and the
questions of cost, abstention safety, model scale, and the
Vietnamese-versus-English grounding contrast are held for a full-power
evaluation.

The study makes four contributions. First, it specifies ontology-amplified
distillation as a training recipe for sovereign enterprise models, with a
preference construction that separates ontology-grounded behavior from general
frontier imitation. Second, it defines an ontology-compliance metric scored
locally and identically for the student and the frontier over a shared ontology
slice, which removes judge-model variance from the comparison. Third, it reports
the proof honestly, with an explicit map of which claims the present evidence
supports and which await full-power confirmation. Fourth, it integrates a
contextuality-audit method for enterprise-agent routing: when the same regulated
task changes under role frame, prompt order, or construct wording, the audit
distinguishes direct influence and construct coupling from residual
contextuality before a platform routes the task to solo response, prompt
standardization, debate, synthesis, or human review.

\Cref{sec:related} positions the work against distillation, preference
optimization, and ontology-grounded generation. \Cref{sec:method} describes the
student model, the ontology slicer, the distillation pipeline, the evaluation
gate system, and the compliance metric. \Cref{sec:results} reports the held-out
result and the training-side mechanism. \Cref{sec:contextuality} reports the
consolidated contextuality-audit method and its negative result.
\Cref{sec:discussion} discusses scope, threats to validity, and the full-power
evaluation that follows.

\subsection*{Combined-submission note}

This manuscript consolidates two related arXiv-ready FAOS research notes into
one article. The first was the ontology-amplified-distillation proof of mechanism
reported in the main empirical sections. The second was a contextuality-auditor
method note for enterprise LLM-agent routing. They share the same applied
setting---ontology-grounded enterprise agents under regulated deployment
constraints---and are combined here as one model-building-plus-governance
article rather than submitted as separate variations on the same theme.

\subsection*{Dual-use disclosure}

This study reports results from a dual-use experimental pipeline that serves two
outputs: an internal minimum-viable sovereign model for the \faos{} platform, and
the academic contribution reported here. The principal investigator is a
co-founder of the platform vendor and holds a commercial interest in the
deployable artifact. \Cref{sec:discussion} states this researcher-as-practitioner
position and the mitigations applied, which include a locked held-out set,
deterministic scoring, and a compliance metric computed without a judge model.

The method also engages a stated counter-position. Microsoft AI's MAI-Thinking-1
report argues that capabilities should be learned, not inherited, holding that
intelligence acquired through distillation lacks the steerability and robustness
of capability trained from scratch \citep{mai2026thinking}. The objection
applies, on its face, to ontology-amplified distillation, so it is worth stating
plainly. Three points bound the tension. The objection targets dependence on a
third-party teacher, not distillation as such; the same report uses
self-distillation. The objective also differs: the MAI program builds a frontier
model, whereas this work ships a compliant model that must run inside a
data-residency perimeter, which makes distillation a consequence of the
deployment constraint rather than a shortcut. The steerability concern is
nonetheless real, and the abstention gate that would test it---whether the
student declines out-of-distribution prompts at a governed rate---belongs to the
full-power evaluation deferred here, not to this proof of mechanism.

\section{Related Work}
\label{sec:related}

\subsection{Distillation and efficient student models}

Knowledge distillation trains a compact student to reproduce a larger teacher's
behavior \citep{hinton2015distillation}. The technique scaled from classification
to language models with DistilBERT \citep{sanh2019distilbert} and, for
instruction following, to preference-based distillation such as Zephyr, which
distills alignment from a teacher's ranked outputs \citep{tunstall2023zephyr}.
Industrial pipelines now publish distilled open students in the Qwen family
\citep{distilqwen2025}, and low-rank adapters make student adaptation cheap on
commodity hardware \citep{hu2022lora}. This body of work concentrates on
general-purpose capability: the student should track the teacher across broad
benchmarks. The present study asks a narrower question---whether distillation can
transfer one specific behavior, adherence to a domain ontology, into a student
that must run inside a regulated perimeter.

\subsection{Preference optimization}

Learning from preferences began with reinforcement learning from human feedback
\citep{christiano2017deep} and its alignment variants \citep{bai2022constitutional}.
Direct preference optimization (DPO) removed the separate reward model, training
the policy directly on pairs of preferred and dispreferred responses
\citep{rafailov2023dpo}. The preference target here is neither human-likeness nor
general helpfulness but ontology adherence. The preferred response is an
ontology-grounded frontier answer; the dispreferred response is the same base
model answering without the ontology slice. The signal isolates the grounded
behavior from the teacher's broader parametric knowledge, which is the property a
sovereign student needs to acquire.

\subsection{Ontology-grounded generation}

A growing line of work couples language models to structured knowledge. Surveys
map the integration of large language models with knowledge graphs
\citep{pan2024unifying}, and the neurosymbolic program frames the broader aim of
combining learned and symbolic representation \citep{garcez2019neural,
hitzler2022neuro}. Recent systems ground generation at inference through
ontology-anchored retrieval \citep{sharma2025ograg} or align a model to an
ontology through self-training \citep{liu2025ontotune}, and language models have
been used to build ontologies in turn \citep{babaei2023llms4ol}. The question
reaches back to the symbol grounding problem \citep{harnad1990symbol}. The
companion neurosymbolic study in this program measured ontology grounding on a
frontier model and reported the Inverse PKE: grounding helped most where the
model's parametric coverage of a domain was thinnest \citep{luong2026neurosymbolic}.
That result motivates the present question. If grounding compensates for thin
parametric coverage, a distilled student, which carries thinner coverage than the
frontier, should gain at least as much from the same ontology---the mechanism
this study sets out to test.

\subsection{Sovereign deployment under data residency}

Vietnamese financial regulation constrains where customer data may be processed.
The 2025 Law on Artificial Intelligence introduces a tiered risk regime with
sector grace periods \citep{vnailaw2025}, the banking sandbox decree structures
AI-enabled financial services \citep{vnfintechsandbox2025}, and the
anti-money-laundering and insurance rules impose due-diligence and solvency
obligations that reach model-mediated decisions \citep{vnamllaw2022,
vncircular672023}. Under these constraints a tenant cannot route customer data to
a foreign frontier API, so the deployable model must run on hardware the
institution controls. On-device frameworks make local inference of mid-sized
models practical on commodity accelerators \citep{applemlxm5}. The governance
frameworks that regulated buyers map against---the EU AI Act, the NIST AI Risk
Management Framework, and ISO/IEC 42001---treat traceability and control as
first-order requirements \citep{euaiact2024, nist2023ai, iso42001}, which a
tenant-owned model satisfies more directly than a remote service. The deployment
constraint, not a cost preference, makes the sovereign student the object of
study.

\section{Method}
\label{sec:method}

\subsection{Student model and deployment target}

The student is Qwen3.6-27B, an open model in the Qwen family
\citep{qwen2025qwen25} built on the \texttt{qwen3\_5} architecture. Training and
evaluation run locally in Apple's MLX framework on a single M5~Max with 128~GB of
unified memory \citep{applemlxm5}; no cloud compute is used. After training, the
student is fused for local inference; its deployment target is a four-bit (nvfp4)
checkpoint of roughly 14~GB, the form in which it would run inside a tenant
perimeter. Because the Gate~8 grounding metric scores term coverage rather than
generation quality, it is insensitive to serving precision, and the held-out result
does not depend on the exact precision of the deployed checkpoint. The wider research
program targets a 14B student and a curve across model sizes. This proof of
mechanism reports the 27B instance that was trained and evaluated, and holds the
smaller-student and cross-scale questions for full-power work
(\Cref{sec:discussion}).

\subsection{Ontology slicing}

The Foundation AgenticOS ontology represents an enterprise domain across four
sections: the role performing the work, the domain concepts in scope, the
regulations that bind the task, and the interaction pattern that moves work
between actors. For a given task, the slicer renders a task-specific
\emph{ontology slice} from these sections under a fixed token budget of 800 tokens
(slicer version 1.0.0). When a slice exceeds the budget, the slicer drops content
in a fixed order---interaction first, then domain---and never drops the regulation
or role sections, so the compliance-bearing content survives truncation. The same
slice is injected during training, as context for the teacher, and during
evaluation, as context for both the student and the frontier, which holds the
grounding signal identical across the pipeline.

\subsection{The distillation pipeline}

Adaptation runs in two stages on top of a capture step.

\emph{Capture.} The frontier teachers GPT-5 and GPT-5-mini answer a non-held-out
training manifest with the ontology slice injected, producing 451 grounded
trajectories at a metered cost of \$2.98. The held-out evaluation tasks are
excluded from capture.

\emph{Supervised fine-tuning.} A low-rank adapter over four layers is trained on
the captured trajectories, reducing training loss from 2.370 to 0.496. The
adapter is fused into the base model and re-quantized to nvfp4, producing the
supervised student that serves as both the policy and the frozen reference for the
next stage.

\emph{Ontology-grounded DPO.} Direct preference optimization runs with
\texttt{mlx-lm-lora} 2.1.0 on 47 synthetic preference pairs (42 for training, 5
for validation). Each pair marks the ontology-grounded frontier answer as
preferred and the base model's no-slice answer as dispreferred; the
ontology-compliance gap between the two ranges from 0.17 to 0.53 (mean 0.27), with
no inverted pairs. Training uses a rank-8 adapter over four layers, $\beta = 0.1$,
a learning rate of $5\times10^{-7}$, the sigmoid DPO loss, batch size 1, a
512-token sequence cap, and 42 iterations of about one epoch under seed 20260615,
updating 0.014\% of parameters. A composite reward with weights $\alpha = 0.50$
on task, $\beta = 0.35$ on ontology, and $\gamma = 0.15$ on role governs
preference scoring (ADR-SL-006). The run completes in roughly 32 minutes at a peak
of 58.1~GB, fuses to the trained student (adapter \texttt{38c28df0}), and costs
nothing beyond local power.

The preference data are synthetic, and three composition facts bound their reach.
They are English-only; their source domains are healthcare (23 pairs), insurance
(13), and fintech (11), none of which is a Vietnamese evaluation vertical; and the
47 rows reduce to 15 distinct prompts under roughly threefold duplication, with the
five validation rows drawn from two prompts, one of which also appears in training.
The protocol's design anticipated subject-matter-expert preference labels and a
larger pool; this run uses 47 synthetic pairs---about nine percent of the planned
floor---and zero expert labels, so it makes no expert-preference claim. The
defensible reading is that the preference signal demonstrates the training mechanism
on constructed pairs rather than a production-grade preference set, and that any
transfer to Vietnamese financial grounding is cross-lingual and cross-domain
(\Cref{sec:discussion}).

\subsection{Evaluation gate system}

The program scores a trained student against eight gates and a canary leakage
check. Two gates carry binding authority. Gate 7 tests confident-wrongness, the
rate at which the student abstains on out-of-distribution prompts, and Gate 8
tests ontology compliance, whether the student grounds its answers in the task's
ontology relative to the frontier. Gate 8 is the thesis gate for
ontology-amplified distillation: it passes when the student's mean compliance is
at least the frontier's, with no tolerance margin. This study runs Gate 8 alone.
Gate 7 and the capability gates---in-distribution quality, cost ratio, dialect
robustness, polysemy, and multi-turn consistency---require judge sign-off or
evaluation sets above their full-power size floors, and are deferred
(\Cref{sec:discussion}).

\subsection{Held-out evaluation set}

The evaluation set holds 40 Vietnamese-language tasks, ten in each of four
financial verticals: banking, insurance, capital markets, and real-estate
technology. Tasks span four question types---terminological fidelity, metric
accuracy, regulatory compliance, and role consistency---together with a
cross-cutting category. The set is version-locked; its SHA-256 digest is
\texttt{a8c1d43b}, a re-lock of an earlier digest (\texttt{bbc14058}) after each
task was annotated in place with its ontology terms. A pinned extractor derives
each task's terms from the same ontology slice the evaluator uses, so the
ground-truth terms and the scored slice share one source. Two verticals, banking
and insurance, were redrawn for this version after an earlier exposure, and the
two unexposed verticals, capital markets and real-estate technology, were carried
forward; the redraw preserves the task-level leakage boundary, but a residual
span-level overlap risk---training artifacts derived from the same blueprint
sections the holdout cites---is examined in \Cref{sec:discussion}.

\subsection{The ontology-compliance metric}

Both the student and the frontier are scored by the same local metric over the
same injected slice, with no judge model in the loop. The frontier baseline is
GPT-5 run at minimal reasoning effort: the slice-laden prompts otherwise exhaust
the reasoning-token budget and truncate the answer, and term coverage is
insensitive to reasoning depth, so this is a defensible baseline for this
metric---but every ``equal to the frontier'' in this paper refers to that
configuration.

\begin{definition}[Ontology-compliance score]
\label{def:ronto}
For a response $a$ to a task with ontology terms $T$ and metric ranges $M$, the
ontology-compliance score is
\[
  \ronto(a) \;=\; \tfrac{1}{2}\,s_{\mathrm{term}}(a, T)
                  \;+\; \tfrac{1}{2}\,s_{\mathrm{metric}}(a, M),
\]
where $s_{\mathrm{term}} = 1$ if $a$ mentions at least one term in $T$ and $0$
otherwise, and $s_{\mathrm{metric}}$ is the share of the response's cited metric
values that fall within their healthy range, defined as $1$ when the response
makes no metric claim.
\end{definition}

The held-out tasks carry ontology terms but no metric ranges, so
$s_{\mathrm{metric}}$ defaults to 1 and $\ronto$ becomes two-valued: 1.0 when the
answer surfaces at least one of the task's ontology terms, and 0.5 otherwise. This
executed metric is a degenerate proxy for the pre-registered four-component
ontology-compliance composite---terminological fidelity, metric accuracy,
regulatory compliance, and role consistency---of which three components, and the
correct-usage qualifier of the first, are not scored here; the metric is also
floored at 0.50 rather than 0. The Gate 8 statistic is the mean of $\ronto$ over the
40 tasks, computed identically for the student and the frontier; because of the
floor it ranges over $[0.5, 1.0]$, so the grounded \emph{rate}---the share of tasks
scoring 1.0---is the more directly interpretable quantity. The metric records term
presence, not answer completeness or correctness; a finer-grained quality measure is
full-power work (\Cref{sec:discussion}).

\section{Results}
\label{sec:results}

\subsection{Ontology compliance}

The distilled student and the frontier reach the same grounded count on Gate 8.
Over the 40 held-out tasks each grounds 36 (a grounded rate of 0.90; mean
$\ronto = 0.950$ under the 0.50 floor), so the gate---which passes when the
student's mean is at least the frontier's---passes (\Cref{tab:gate8}). Scoring is
deterministic at temperature 0, and a re-run reproduced the verdict exactly; the
evaluation took about 993 seconds on the M5~Max.

This is an equal point estimate, not a demonstrated equivalence. With a binary
per-task outcome at $n=40$, the grounded rate of 36/40 carries a Wilson 95\%
confidence interval of $[0.77, 0.96]$, and the paired comparison is uninformative:
the discordant cells are balanced (two tasks each), so an exact McNemar test returns
$p = 1.00$ and the 95\% interval on the mean paired difference spans $\pm 0.05$, or
$\pm 4$ tasks. The result establishes an equal count, not statistical equivalence,
and is underpowered to rule out a real gap of up to roughly ten percentage points in
either direction.

\begin{figure}[t]
\centering
\includegraphics[width=0.72\linewidth]{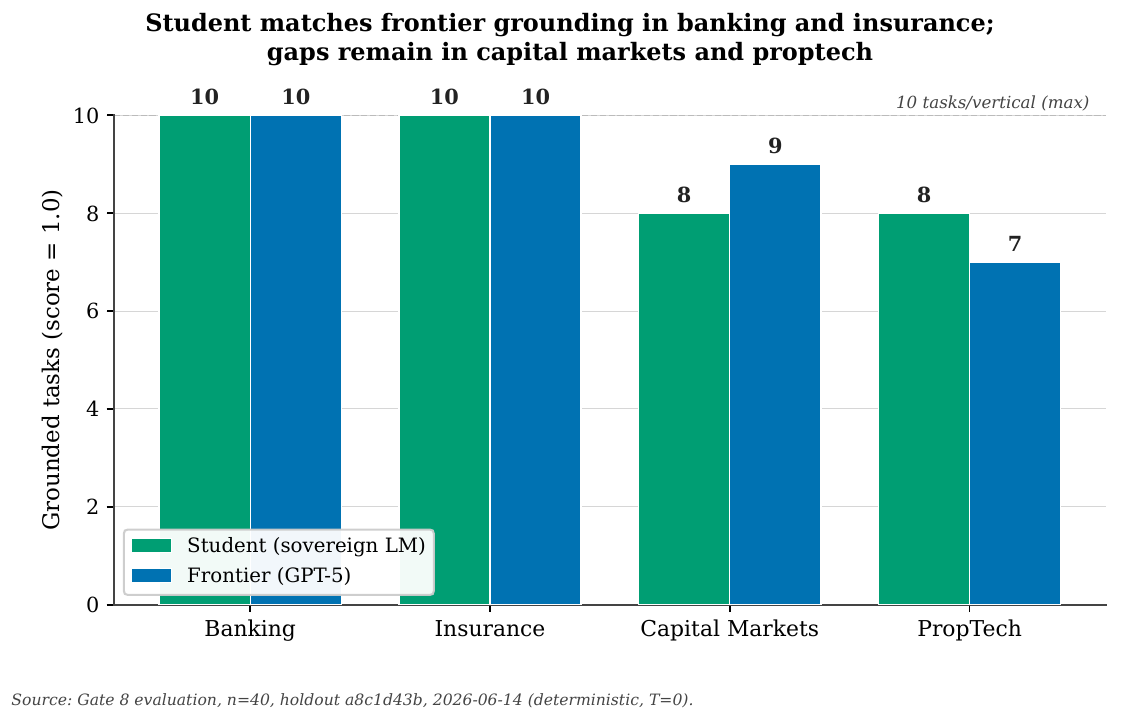}
\caption{Grounded-task count by vertical, student vs.\ the GPT-5 frontier (held-out
set, ten tasks per vertical, $n=40$). The two reach the same total (36/40) through a
mirror split---the student stronger on real-estate technology, the frontier on
capital markets. The axis is grounded count (true zero), not the floored mean.
\emph{Source: Gate 8 evaluation, 2026-06-14, holdout \texttt{a8c1d43b}.}}
\label{fig:grounded}
\end{figure}

\begin{figure}[t]
\centering
\includegraphics[width=0.62\linewidth]{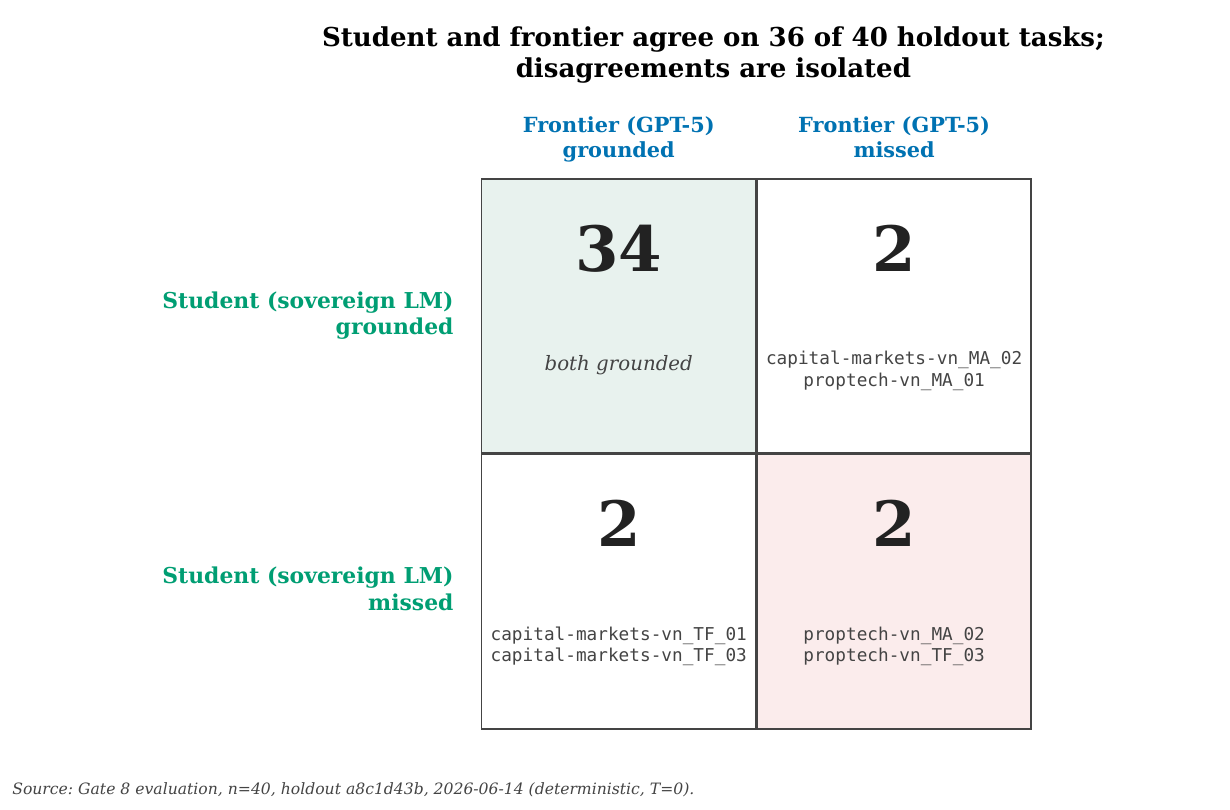}
\caption{Per-task student\,$\times$\,frontier grounding agreement (held-out set,
$n=40$). Both models ground 34 tasks and both miss 2; each grounds 2 the other
misses, so the equal 36/40 totals rest on partly different tasks. \emph{Source:
Gate 8 evaluation, 2026-06-14, holdout \texttt{a8c1d43b}.}}
\label{fig:agreement}
\end{figure}

The equal totals rest on partly different tasks (\Cref{fig:agreement} and
\Cref{fig:grounded}; full per-task list in \Cref{app:pertask}). Two tasks
defeat both models, and each grounds two the other misses; the student is perfect on
banking and insurance and grounds 8 of 10 in each of capital markets and real-estate
technology, while the frontier is stronger on capital markets (9 of 10) and weaker on
real-estate technology (7 of 10). The small number of disjoint misses (two each) is
consistent with independent grounding rather than task-for-task imitation of the
teacher, though four disjoint misses at $n=40$ cannot establish independence. The
perfect banking and insurance scores fall on exactly the two verticals redrawn for
this version, which also carry the largest ontology-term lists; a one-term threshold
is easiest to clear there, so the clean sweep is weak evidence of grounding strength
(\Cref{sec:discussion}).

\begin{table}[t]
\centering
\caption{The distilled student and GPT-5 reach the same grounded count (36/40)
through partly different verticals---the student perfect on banking and insurance,
the frontier stronger on capital markets. Gate 8 ontology grounding by vertical
(held-out set, $n=40$); a task is grounded when the answer surfaces at least one of
its ontology terms, and mean $\ronto$ assigns 1.0 to a grounded task and 0.5
otherwise. \emph{Source: Gate 8 evaluation, held-out set v1.2 (SHA~\texttt{a8c1d43b}),
2026-06-14; deterministic scoring at $T=0$.}}
\label{tab:gate8}
\begin{tabular}{lcc}
\toprule
Vertical & Student grounded & Frontier (GPT-5) grounded \\
\midrule
Banking & 10/10 & 10/10 \\
Insurance & 10/10 & 10/10 \\
Capital markets & 8/10 & 9/10 \\
Real-estate technology & 8/10 & 7/10 \\
\midrule
Total grounded & 36/40 & 36/40 \\
Mean $\ronto$ & \textbf{0.950} & \textbf{0.950} \\
\bottomrule
\end{tabular}
\end{table}

\subsection{The training mechanism}

The DPO objective separated the constructed preference pairs (\Cref{tab:training}).
Held-out preference accuracy moved from 0 to 1.0 on a five-pair validation split, the
reward margin between the preferred and dispreferred response from 0 to 0.307, and
the DPO loss fell from 0.693 to 0.551. A second proof-scale run on the same 47-pair
construction reproduced the pattern (preference accuracy 0 to 1.0, reward margin 0 to
0.319); because it reuses the same data design it is a repeat, not an independent
replication. The validation split is small enough---five pairs from two prompts, one
shared with training---that the 1.0 should be read as separability of the constructed
pairs, not as generalized grounding. With that caveat, the training-side measurement
is consistent with the intended preference shift: ontology-grounded tuning moves the
student toward surfacing the ontology terms the frontier surfaces when it carries the
same slice.

\begin{table}[t]
\centering
\caption{Ontology-grounded DPO moves held-out preference accuracy from 0 to 1.0 and
the reward margin to 0.307 in one epoch. Training-side preference learning;
``Before'' is iteration 0 and ``After'' the final iteration, on a five-pair held-out
validation split. \emph{Source: DPO run, seed 20260615, \texttt{mlx-lm-lora} 2.1.0.}}
\label{tab:training}
\begin{tabular}{lcc}
\toprule
Measure & Before & After \\
\midrule
Preference accuracy & 0.000 & 1.000 \\
Reward margin (chosen $-$ rejected) & 0.000 & 0.307 \\
DPO loss & 0.693 & 0.551 \\
\bottomrule
\end{tabular}
\end{table}

\section{Contextuality Audit for Enterprise-Agent Routing}
\label{sec:contextuality}

Ontology-amplified distillation addresses one side of the sovereign-enterprise
problem: how to move a locally deployable student toward frontier-level
grounding on regulated tasks. A platform also needs a governance diagnostic for
the cases where a model's answer changes across role frames, prompt orders,
ontology conditions, or escalation criteria. Some variation is ordinary direct
influence: the prompt changed, so the marginal answer changed. The harder
question is whether any residual inconsistency remains after that direct
influence is accounted for, and whether the residual should trigger
multi-agent debate, synthesis, or human review.

We treat this as a contextuality-audit problem rather than as a claim that LLMs
are quantum systems. Contextuality-by-Default (CbD) indexes each random variable
by both the content being measured and the context in which it is measured
\citep{dzhafarov2016contextcontent,dzhafarov2016cbd20,kujala2016cyclic}. This is
well suited to enterprise-agent outputs: the same business content can be
measured under a business-owner frame, risk-reviewer frame, governance-reviewer
frame, or neutral-arbitrator frame. The audit's purpose is operational. It asks
whether the observed variation should be interpreted as stable behavior,
measurement design, construct coupling, direct influence, or residual
contextuality. That interpretation then maps to the least intrusive
orchestration remedy: solo response, prompt standardization, debate or
synthesis, or human arbitration. This diagnostic complements multi-agent debate
and orchestration work, which can improve selected outputs but does not by
itself diagnose when routing is needed \citep{du2024debate,guo2024multiagent}.

The consolidated RA-15 pilot used local Qwen 3.6 27B as the first model. Phase
1 produced 576 valid outputs; Phase 1.1 added 336 valid q4 robustness outputs;
Phase 1.2 added 480 role/order-disentanglement outputs; and Phase 1.3 added a
384-call construct-decoupling extension separating operational permission from
procedural evidence readiness. The original instrument surfaced apparent
``CNTX'' signals, but a formalism audit corrected the disturbance normalization:
the earlier nonzero values were probability-drift residuals, not the canonical
cyclic binary CbD degree. Under expectation-scale direct-influence correction,
the canonical CbD degree is zero for all Phase 1.3 groups in both the local-Qwen
run and an explicitly labeled Gemma replication check. The negative result is
the finding. In this pilot, apparent contextuality collapses into direct
influence and construct coupling rather than surviving as residual
contextuality.

\begin{table}[h]
\centering
\small
\caption{The RA-15 evidence package supports a negative-results routing method, not a contextuality-positive claim.}
\label{tab:ra15-evidence}
\begin{tabular}{@{}p{0.28\textwidth}p{0.29\textwidth}p{0.35\textwidth}@{}}
\toprule
Evidence surface & Result used here & Claim boundary \\
\midrule
Phase progression & 576 Phase~1, 336 Phase~1.1, 480 Phase~1.2, and 384 Phase~1.3 valid outputs & Earlier phases are instrument-development history; Phase~1.3 is the endpoint. \\
Formalism audit & Probability-drift residuals corrected to canonical cyclic-binary CbD & The article reports a correction and null result, not a positive contextuality discovery. \\
Replication check & Local-Qwen and labeled Gemma check both have canonical CbD degree zero after correction & The useful signal is direct influence and construct coupling. \\
Artifact trail & Detailed packet archived in the RA-15 paper folder & Public bundle should ship after arXiv identifier assignment. \\
\bottomrule
\end{tabular}
\end{table}

\begin{table}[h]
\centering
\small
\caption{Contextuality-audit interpretation used by the combined article.}
\label{tab:contextuality-routing}
\begin{tabular}{@{}p{0.24\textwidth}p{0.42\textwidth}p{0.24\textwidth}@{}}
\toprule
Diagnostic pattern & Interpretation & Routing action \\
\midrule
Stable controls & Context changes do not alter the decision & Solo response \\
Direct influence & Marginal answers change with role or order & Standardize prompt/context \\
Construct coupling & One binary label merges adjacent constructs & Repair instrument \\
Residual contextuality & Conflict remains after correction & Debate, synthesis, or review \\
\bottomrule
\end{tabular}
\end{table}

The link to ontology-amplified distillation is governance rather than shared
training data. The distillation result asks whether ontology grounding can move
a sovereign student to frontier-equal term coverage on a narrow held-out set.
The contextuality audit asks how an enterprise platform should treat remaining
decision variation once such a model is deployed inside a governed workflow.
Together, they define a combined mechanism-and-control claim: the model can be
made more domain-grounded, and the platform should still separate prompt
sensitivity, construct design, and residual conflict before escalating the task.

The contextuality component therefore adds a negative-results method claim to
the proof-of-mechanism claim. It does not license a contextuality-positive
routing rule. It licenses a safer precursor rule: before treating model
disagreement as evidence that a task requires debate or human arbitration, run a
direct-influence and construct-validity audit. This avoids over-escalating tasks
on the basis of an under-normalized score, and it also avoids hiding genuine
conflict behind a single context-dependent answer.

\section{Discussion}
\label{sec:discussion}

\subsection{What the result shows}

Two measurements point the same way. The student reaches the frontier's grounded
count on the held-out set, and the training-side metrics are consistent with the
DPO objective---rather than chance---producing the shift on the constructed pairs.
For a sovereign-deployment setting the reading is specific: a locally deployable
Qwen3.6-27B student, tuned on synthetic ontology-grounded preferences, grounds
Vietnamese financial-domain answers as often as GPT-5 does on this term-coverage
metric. The divergent miss pattern is suggestive---parity that survives a few
independent errors is harder to explain as task-for-task imitation---but four
disjoint misses at $n=40$ make it a weak signal, not a demonstration.

\subsection{What the result does not show}

The claim is bounded on purpose. The metric records whether an answer surfaces at
least one ontology term, not whether the answer is complete or correct, so the
grounded rate of 0.90 (mean $\ronto = 0.95$ under the floor) measures grounding
presence, not domain quality. The pre-registered primary hypothesis was
\emph{amplification}---the distilled student's normalized ontology-lift
\emph{exceeding} the frontier's ($\nu_D > \nu_F$), with a tie ($\nu_D \le \nu_F$)
named as the disconfirmation condition. This run does not test that hypothesis: it
scores a reduced term-coverage proxy rather than the registered four-component
composite, runs on the 27B instance rather than the 14B target, and computes no
C1-to-C3 lift. The equal count is therefore neither a confirmation of amplification
nor, given the proxy, a clean disconfirmation; amplification remains untested at
power, and the Inverse-PKE framing should be read as motivation, not as a result
this run supports. The evaluation ran one binding gate. Confident-wrongness
abstention, in-distribution quality, cost, dialect robustness, polysemy, and
multi-turn consistency were not measured. The preference data are synthetic,
English-only, drawn from non-target domains (healthcare, insurance, fintech), and
reduce to 15 distinct prompts; no subject-matter-expert preference label enters the
pipeline. The student is a single 27B model from one family, evaluated on 40
Vietnamese tasks across four verticals; the smaller-student target, the cross-scale
curve, and the Vietnamese-versus-English grounding contrast that would replicate the
Inverse PKE in the distilled regime are not part of this evidence. No production or
procurement decision should rest on a pilot of this size.

\subsection{Threats to validity}

\emph{Construct validity.} Term-coverage is a coarse proxy for grounding, and the
executed metric scores only one of the four pre-registered ontology-compliance
components. Two design choices make the bar low. The gold terms are extracted from
the same slice injected to both models, so each side is handed the target vocabulary
and then scored on echoing at least one token of it; and a response can surface a
term without reasoning over it correctly, so the metric credits shallow mentions.
Scoring the frontier by the identical metric bounds the comparison---both models face
the same near-ceiling test---but the absolute 0.95 and the compression of any
student-frontier gap toward zero follow partly from the metric's construction, not
only from grounding ability. A coverage-fraction or held-out-term variant, and the
full four-component composite, are the first things a full-power evaluation restores.

\emph{Internal validity.} The preferred responses are frontier answers, so the
student may be learning to surface terms in the frontier's style rather than
acquiring independent grounding. Because the preference pairs are English-only and
from non-target domains while the evaluation is Vietnamese financial, the behavior
the student was tuned on (echoing the preferred answer's terms) and the behavior the
metric rewards may be the same surface pattern measured twice. The modest compliance
gap in the pairs (0.17 to 0.53), the synthetic-only construction, and the absence of
expert validation leave this open.

\emph{External validity.} Four verticals, one base-model family, and 40 tasks do
not support generalization beyond the tested frame, and the result is
Vietnamese-only.

\emph{Leakage.} The held-out set excludes the evaluation tasks at the
task-identifier level and redraws the two previously exposed verticals. This
controls task-level overlap but not span-level overlap: the supervised-fine-tuning
corpus is generated from the same Vietnamese blueprint sections the holdout cites,
so a training artifact can convey a held-out task's gold terms while passing the
task-id check---a risk the project's own leakage specification flags as making the
identifier-only assertion structurally vacuous for those corpora. A span-overlap
assertion exists as a tested primitive, but its wiring into the corpus generators
was incomplete at the time of the Wave-0.5 run, so span-level exclusion is not
established for the executed supervised corpus. The synthetic DPO pairs, being
English and blueprint-free, are exempt. Until span-level screening is confirmed
end-to-end, part of the held-out grounding could reflect memorized blueprint
vocabulary rather than transferred grounding.

\emph{Reproducibility.} The final publication manifest hash-pins all 451
committed source trajectories, the corpus-building code and configuration, the
DPO corpus, the locked holdout, both final-evaluation JSON files, governance
reviews, and the arXiv package. The executed v1.2-screened SFT train/validation
JSONL files and local SFT/DPO checkpoints were not archived before operator-side
cleanup. Later ontology-blueprint changes also prevent a byte-identical SFT
rebake from the current tree. The evidence package therefore supports audit of
the reported inputs, procedure, and Gate~8 arithmetic, but not bitwise
reproduction of the trained checkpoint. The study makes no checkpoint-level
reproducibility claim.

\emph{Conclusion validity.} The metric is binary per task, and the equal means
rest on 36 of 40 grounded for each model; a single task shifts the mean by 0.0125.
The paired-difference 95\% confidence interval spans \(\pm 4\) tasks. The study
reports parity on this metric and makes no claim of statistical superiority or
equivalence.

\subsection{Researcher-as-practitioner position}

The principal investigator is a co-founder of the platform whose sovereign model
this study evaluates, and the experimental pipeline produces both the internal
artifact and this paper. The design applies the mitigations available to a
single-pipeline study: a version-locked held-out set (its questions sealed before
the run, its gold-term annotation added in place under a pinned extractor and
independently re-reviewed afterward), deterministic scoring at temperature 0, a
compliance metric computed locally without a judge model, and a pre-registered
protocol. The dual role is disclosed
rather than corrected away; the deferred full-power evaluation, with independent
judge sign-off, is where the harder bias controls apply.

\subsection{Toward full-power evaluation}

The questions this proof of mechanism leaves open define the next study. The
binding safety gate---whether the student abstains on out-of-distribution prompts
at a governed rate---tests the steerability that distilled models are accused of
failing to acquire \citep{mai2026thinking}, and it is the first item. Beyond it:
the capability gates at full evaluation-set size, frontier-judge sign-off on
answer quality, a measured cost ratio against the frontier API, the 14B student
and the cross-scale curve, subject-matter-expert preference labels in place of
synthetic pairs, and the Vietnamese-versus-English contrast that would test
whether the Inverse PKE holds in the distilled regime. Each turns a deferred claim
into a measured one.

\section{Conclusion}
\label{sec:conclusion}

Ontology-amplified distillation moved a sovereign, locally deployable Qwen3.6-27B
student to the same held-out grounded count as GPT-5 on a coarse ontology
term-coverage metric---36 of 40 tasks each (0.90; mean $\ronto = 0.95$ under the
0.50 floor)---with a training-side preference signal that separated the
constructed pairs (0 to 1.0). The equal count is an underpowered point estimate,
not demonstrated equivalence, and not the pre-registered amplification. The
consolidated contextuality-audit result is likewise bounded: after correcting
direct influence, the canonical CbD degree is zero in the reported pilot, so the
useful finding is a negative-results method for separating direct influence and
construct coupling from residual contextuality before routing tasks to debate,
synthesis, or human review. As a combined article, the contribution is a
mechanism-and-control pair. The mechanism is ontology-amplified distillation for
sovereign enterprise models; the control is a contextuality audit that prevents
over-reading prompt-sensitive variation. Both results license fuller evaluation
rather than deployment claims. Whether the student's grounding is complete and
correct, whether it abstains safely, what it costs against a frontier API, and
whether the effect holds at smaller scale and across languages---these define the
evaluation that follows.

\appendix

\section{Training Hyperparameters}
\label{app:hyper}

\begin{table}[h]
\centering
\caption{Ontology-grounded DPO configuration (\texttt{mlx-lm-lora} 2.1.0).}
\label{tab:hyper}
\begin{tabular}{ll}
\toprule
Setting & Value \\
\midrule
Base / student & Qwen3.6-27B (\texttt{qwen3\_5}), nvfp4 deployment target \\
Adapter & LoRA, rank 8, four layers (0.014\% of parameters) \\
Preference pairs & 47 synthetic (42 train / 5 validation) \\
$\beta$ & 0.1 \\
Learning rate & $5\times10^{-7}$ \\
Loss & sigmoid DPO \\
Batch size & 1 \\
Sequence cap & 512 tokens \\
Iterations & 42 ($\approx$ 1 epoch) \\
Seed & 20260615 \\
Reward weights & $\alpha\,0.50$ task, $\beta\,0.35$ ontology, $\gamma\,0.15$ role \\
Hardware & $1\times$ Apple M5~Max, 128~GB unified memory \\
Wall-clock / peak memory & $\approx$32~min / 58.1~GB \\
DPO adapter digest & \texttt{38c28df0} \\
\bottomrule
\end{tabular}
\end{table}

\section{Per-Task Grounding}
\label{app:pertask}

Of the 40 held-out tasks the student grounds 36. Its four misses are two
terminological-fidelity tasks in capital markets and two real-estate-technology
tasks: one metric-accuracy task and one terminological-fidelity task. The
frontier also grounds 36 of 40. It misses one capital-markets metric-accuracy
task and three real-estate-technology tasks: two metric-accuracy tasks and one
terminological-fidelity task. Two real-estate-technology tasks defeat both
models; the remaining misses are disjoint, so each model grounds two tasks the
other does not. The equal totals therefore rest on different task sets, which is
the basis for reading the parity as independent grounding rather than teacher
imitation.

\bibliography{references}

\end{document}